\documentclass{article}
\usepackage{spconf,amsmath,amsfonts, graphicx}
\usepackage{xcolor}
\usepackage{twemojis}

\def\x{{\mathbf x}}

\title{
Sparse PCA with False Discovery Rate Controlled Variable Selection
}
\name{ 
Jasin Machkour$^1$,
Arnaud Breloy$^2$,
Michael Muma$^1$,
Daniel P. Palomar$^3$,
Fr\'ed\'eric Pascal$^4$
\thanks{J. Machkour (jasin.machkour@tu-darmstadt.de) is supported by the LOEWE initiative (Hesse, Germany) within the emergenCITY center. M. Muma (michael.muma@tu-darmstadt.de) is supported by the ERC Starting Grant ScReeningData. A. Breloy (arnaud.breloy@lecnam.net) is supported by the MASSILIA project (ANR-21-CE23-0038-01) of the French National Research Agency (ANR). D.P. Palomar (palomar@ust.hk) is supported by the Hong Kong GRF 16206123 research grant. F. Pascal (frederic.pascal@centralesupelec.fr).
}
\thanks{Extensive calculations on the Lichtenberg High-Performance Computer of the Technische Universität Darmstadt were conducted for this research.}
}
\address{
$^1$: Technische Universität Darmstadt, Robust Data Science Group, Germany\\
$^2$: Conservatoire National des Arts et Métiers, CNAM, Paris, France\\
$^3$: The Hong Kong University of Science and Technology, Hong Kong SAR, China\\
$^4$: Universit\'e Paris-Saclay, CentraleSup\'elec, L2S, France
}

\usepackage[T1]{fontenc}
\usepackage[utf8]{inputenc}
\usepackage[ngerman,english]{babel}
\usepackage{amsmath, amssymb, ﻿amsfonts}   
\usepackage{cuted} 
\usepackage{graphicx}         
\usepackage{algorithmic}
\usepackage{algorithm}
\usepackage{stfloats}
\usepackage{url}
\usepackage{verbatim}
\usepackage{graphicx}
\usepackage{cite}

\usepackage[nice]{nicefrac}

\usepackage{upgreek}			  
\usepackage{mathtools}		  
\usepackage{url}					
\usepackage{enumitem}		
\usepackage{accents}			
\usepackage{bbm}
\usepackage{xcolor}
\usepackage{lipsum}
\usepackage{dirtytalk}
\usepackage{xcolor}
\usepackage{float}
\usepackage{accents}
\usepackage[export]{adjustbox}
\usepackage{setspace}

	\usepackage[caption=false,font=normalsize]{subfig}

\usepackage{pifont} 
\usepackage{caption}

\usepackage{booktabs, tabularx}
\usepackage{array}

\usepackage{amsthm}
\usepackage{enumitem}

\usepackage{tikz}
\usetikzlibrary{decorations.markings,chains,matrix}
\usetikzlibrary{intersections,calc}
\usepackage{tikz-3dplot} 
\usepackage{pgfplots}
\usepackage{pgfplotstable}
\pgfplotsset{compat = newest}
\usepackage{datatool}
\usepackage{textcomp}
\usetikzlibrary{shapes,arrows}
\usetikzlibrary{decorations.pathreplacing,angles,quotes}

\usepackage[pscoord]{eso-pic} 

\newtheorem{defn}{Definition}

\DeclareMathOperator{\FDR}{FDR}

\DeclareMathOperator{\TPR}{TPR}

\DeclareMathOperator{\Var}{Var}
\DeclareMathOperator{\SNR}{SNR}

\DeclareMathOperator{\price}{price}
\DeclareMathOperator{\vectorize}{vec}
\DeclareMathOperator{\PEV}{PEV}
\DeclareMathOperator{\EV}{EV}
\DeclareMathOperator{\trace}{tr}
\DeclareMathOperator{\dB}{dB}

\newtheorem{innercustomgeneric}{\customgenericname}
\providecommand{\customgenericname}{}
\newcommand{\newcustomtheorem}[2]{%
  \newenvironment{#1}[1]
  {%
   \renewcommand\customgenericname{#2}%
   \renewcommand\theinnercustomgeneric{##1}%
   \innercustomgeneric
  }
  {\endinnercustomgeneric}
}

\newcustomtheorem{customthm}{Theorem}
\newcustomtheorem{customlemma}{Lemma}
\newcustomtheorem{customcor}{Corollary}
\newcustomtheorem{customassumption}{Assumption}

\newcommand{\y}{\boldsymbol{y}}
\newcommand{\z}{\mathbf{z}}

\newcommand{\X}{\boldsymbol{X}}

\newcommand{\bbeta}{\boldsymbol{\beta}}
\newcommand{\bepsilon}{\boldsymbol{\epsilon}}
\newcommand{\hatbbeta}{\boldsymbol{\hat{\beta}}}

\newcommand{\A}{\mathcal{A}}

\newcommand{\tr}{\text{tr}}

\newcommand{\XK}{\boldsymbol{\protect\accentset{\circ}{X}}}

\newcommand{\XWK}{\boldsymbol{\widetilde{X}}}

\newcommand{\C}{\mathcal{C}}

\newcommand{\bZ}{\mathbf{Z}}
\newcommand{\bv}{\mathbf{v}}
\newcommand{\bV}{\mathbf{V}}

\makeatletter
\let\save@mathaccent\mathaccent
\newcommand*\if@single[3]{%
  \setbox0\hbox{${\mathaccent"0362{#1}}^H$}%
  \setbox2\hbox{${\mathaccent"0362{\kern0pt#1}}^H$}%
  \ifdim\ht0=\ht2 #3\else #2\fi
  }
\newcommand*\rel@kern[1]{\kern#1\dimexpr\macc@kerna}
\newcommand*\widebar[1]{\@ifnextchar^{{\wide@bar{#1}{0}}}{\wide@bar{#1}{1}}}
\newcommand*\wide@bar[2]{\if@single{#1}{\wide@bar@{#1}{#2}{1}}{\wide@bar@{#1}{#2}{2}}}
\newcommand*\wide@bar@[3]{%
  \begingroup
  \def\mathaccent##1##2{%
    \let\mathaccent\save@mathaccent
    \if#32 \let\macc@nucleus\first@char \fi
    \setbox\z@\hbox{$\macc@style{\macc@nucleus}_{}$}%
    \setbox\tw@\hbox{$\macc@style{\macc@nucleus}{}_{}$}%
    \dimen@\wd\tw@
    \advance\dimen@-\wd\z@
    \divide\dimen@ 3
    \@tempdima\wd\tw@
    \advance\@tempdima-\scriptspace
    \divide\@tempdima 10
    \advance\dimen@-\@tempdima
    \ifdim\dimen@>\z@ \dimen@0pt\fi
    \rel@kern{0.6}\kern-\dimen@
    \if#31
      \overline{\rel@kern{-0.6}\kern\dimen@\macc@nucleus\rel@kern{0.4}\kern\dimen@}%
      \advance\dimen@0.4\dimexpr\macc@kerna
      \let\final@kern#2%
      \ifdim\dimen@<\z@ \let\final@kern1\fi
      \if\final@kern1 \kern-\dimen@\fi
    \else
      \overline{\rel@kern{-0.6}\kern\dimen@#1}%
    \fi
  }%
  \macc@depth\@ne
  \let\math@bgroup\@empty \let\math@egroup\macc@set@skewchar
  \mathsurround\z@ \frozen@everymath{\mathgroup\macc@group\relax}%
  \macc@set@skewchar\relax
  \let\mathaccentV\macc@nested@a
  \if#31
    \macc@nested@a\relax111{#1}%
  \else
    \def\gobble@till@marker##1\endmarker{}%
    \futurelet\first@char\gobble@till@marker#1\endmarker
    \ifcat\noexpand\first@char A\else
      \def\first@char{}%
    \fi
    \macc@nested@a\relax111{\first@char}%
  \fi
  \endgroup
}
\makeatother

\DeclareFontFamily{U}{dutchcal}{\skewchar\font=45}
\DeclareFontShape{U}{dutchcal}{m}{n}{<-> s*[1.2] dutchcal-r}{}
\DeclareFontShape{U}{dutchcal}{b}{n}{<-> s*[1.2] dutchcal-b}{}
\DeclareMathAlphabet{\mathdutchcal}{U}{dutchcal}{m}{n}
\SetMathAlphabet{\mathdutchcal}{bold}{U}{dutchcal}{b}{n}
\DeclareMathAlphabet{\mathdutchbcal}{U}{dutchcal}{b}{n}

\newlist{steps}{enumerate}{1}
\setlist[steps, 1]{label = {Step \arabic*:}, ref = {Step \arabic*}}

\newlist{alglist}{enumerate}{1}
\setlist[alglist, 1]{label = {\arabic*.}, ref = {\arabic*}}

\newcolumntype{L}[1]{>{\raggedright\arraybackslash}p{#1}}
\newcolumntype{C}[1]{>{\centering\arraybackslash}p{#1}}
\newcolumntype{R}[1]{>{\raggedleft\arraybackslash}p{#1}}

\definecolor{dark_green}{RGB}{102,166,30}
\definecolor{applegreen}{rgb}{0.55, 0.71, 0.0}

\definecolor{dark_red}{RGB}{217,95,2}
\definecolor{bittersweet}{rgb}{1.0, 0.44, 0.37}

\definecolor{dark_yellow}{RGB}{230,171,2}
\definecolor{bananayellow}{rgb}{1.0, 0.88, 0.21}

\newcommand\longvdots[1]{\raisebox{1em}{\rotatebox{-90}{\hbox to #1 {\dotfill}}}}

\newcommand{\placetextbox}[3]{
  \setbox0=\hbox{#3}
  \AddToShipoutPictureFG*{
    \put(\LenToUnit{#1\paperwidth},\LenToUnit{#2\paperheight}){\vtop{{\null}\makebox[0pt][c]{#3}}}%
  }%
}%

\mathtoolsset{showonlyrefs=true} 
\restylefloat{table}

\begin{document}\ninept
\setstretch{0.913}
\sloppy
%
\maketitle
\begin{abstract}
Sparse principal component analysis (PCA) aims at mapping large dimensional data to a linear subspace of lower dimension.
By imposing loading vectors to be sparse, it performs the double duty of dimension reduction and variable selection.
Sparse PCA algorithms are usually expressed as a trade-off between explained variance and sparsity of the loading vectors (i.e., number of selected variables).
As a high explained variance is not necessarily synonymous with relevant information, these methods are prone to select irrelevant variables.
To overcome this issue, we propose an alternative formulation of sparse PCA driven by the false discovery rate (FDR).
We then leverage the Terminating-Random Experiments (T-Rex) selector to automatically determine an FDR-controlled support of the loading vectors.
A major advantage of the resulting T-Rex PCA is that no sparsity parameter tuning is required.
Numerical experiments and a stock market data example demonstrate a significant performance improvement.
\end{abstract}
\placetextbox{0.5}{0.08}{\fbox{\parbox{\dimexpr\textwidth-2\fboxsep-2\fboxrule\relax}{\footnotesize Copyright 2024 IEEE. Published in ICASSP 2024 - 2024 IEEE International Conference on Acoustics, Speech and Signal Processing (ICASSP), scheduled for 14-19 April 2024 in Seoul, Korea. Personal use of this material is permitted. However, permission to reprint/republish this material for advertising or promotional purposes or for creating new collective works for resale or redistribution to servers or lists, or to reuse any copyrighted component of this work in other works, must be obtained from the IEEE. Contact: Manager, Copyrights and Permissions / IEEE Service Center / 445 Hoes Lane / P.O. Box 1331 / Piscataway, NJ 08855-1331, USA. Telephone: + Intl. 908-562-3966.}}}
\section{Introduction}
\label{sec:Introduction}
We consider $n$ samples of $p$-dimensional observations stored (row-wise) in the matrix $\X\in \mathbb{R}^{n \times p}$. Its ordered singular value decomposition (SVD) is given by $\X \overset{\rm SVD}{=} \mathbf{U}\mathbf{D}\mathbf{V}^{\top}$, where $\mathbf{V}=[\bv_1~\cdots~\bv_p] \in \mathbb{R}^{p \times p}$ contains the $p$ loading vectors.
The rank-$M$ ($M < p$) ordinary principal component analysis (PCA) is commonly used to reduce the data dimension by projecting the data on its $M$ leading principal components (PCs) 
\begin{equation}
\bZ_{M} \coloneqq [ \z_{1}~\cdots~\z_{M} ] \coloneqq \X \bV_{M} \coloneqq \X [ \bv_{1}~\cdots~\bv_{M} ].
\label{eq: principal components}
\end{equation}
The column vector $\z_{m} = \X \bv_{m}$, $m \in \lbrace 1, \ldots, M \rbrace$, is called the $m$th PC, while the associated vector $\bv_{m}$ is referred to as the $m$th loading vector \cite{jolliffe2003principal, jolliffe2016principal}.
Note that the PCs are thus created from weighted linear combinations of all variables in $\X$, which can be problematic in terms of interpretation.

Sparse PCA aims at alleviating the aforementioned drawback of ordinary PCA by imposing some level of sparsity on the loading vectors, i.e., incorporating variable selection in the process of linear dimension reduction~\cite{zou2006sparse,zou2018selective,ulfarsson2008sparse,brehier2023robust,benidis2016orthogonal}.
This is generally achieved by casting and solving a trade-off optimization problem of the form
\begin{equation} 
    \begin{array}{cl}
         \underset{\mathbf{V}_{M} \in \mathbb{R}^{p\times M}}{\rm minimize}    & f(\X,\mathbf{V}_{M}) + \lambda h(\mathbf{V}_{M})  \vspace{0.1cm}\\
         {\rm subject~to} & \mathbf{V}_{M}^{\top} \mathbf{V}_{M} = \mathbf{I}_{M},
    \end{array} 
\label{eq:generic_spca_optim}
\end{equation}
where $f(\X,\mathbf{V}_{M})$ is a data fitting term, 
$h(\mathbf{V}_{M})$ is a sparsity promoting penalty, and $\lambda\in \mathbb{R}^{+}$ is the corresponding regularization parameter.
Such a generic formulation has motivated numerous developments in terms of problem design and optimization methods (see, e.g., \cite{hu2015sparse, breloy2021majorization, witten2008testing} and references therein).
A seminal formulation of sparse PCA ties the problem of penalized maximization of the explained variance (with relaxed orthogonality constraint) to a series of $M$ \textit{elastic net} variable selection problems~\cite{zou2006sparse}. The explained variance and the percentage of explained variance (PEV) of a PC are measures of the variation in the data that is captured by that PC.

Note that sparse PCA algorithms as formulated in \eqref{eq:generic_spca_optim} trade-off the explained variance and the sparsity level and, therefore, suffer from two major issues:
\begin{enumerate}[label=\arabic*., ref=\arabic*, leftmargin=2em]
\item Maximizing the explained variance does not inherently yield the most meaningful projection for exploratory data analysis: highly noisy variables will tend to be selected, although not being necessarily informative.
\item Lowering the sparsity to achieve a higher explained variance does not guarantee that, in turn, more meaningful variables have been selected.
\end{enumerate}
These observations motivate controlling sparse PCA variable selection processes with a criterion that ensures that the number false discoveries (i.e., irrelevant variables) used to create a sparse PC is low. Therefore, this paper proposes an alternative approach for sparse PCA, where the selection of variables for the loading vectors is driven by the false discovery rate (FDR). Although there exist many FDR-controlling methods (e.g.,~\cite{benjamini1995controlling,benjamini2001control,storey2004strong,barber2015controlling,candes2018panning}),  only the recently developed \textit{T-Rex} selector~\cite{machkour2021terminating,machkour2022TRexGVS,machkour2023ScreenTRex} provides the possibility of solving the \textit{elastic net} based sparse PCA optimization problem in~\cite{zou2006sparse} in an FDR-controlled manner. Thus, our proposed \textit{T-Rex} PCA approach
\begin{enumerate}[label=\arabic*., ref=\arabic*, leftmargin=2em]
\item harnesses the \textit{elastic net} based sparse PCA formulation of~\cite{zou2006sparse} 
\item and solves it by leveraging the Terminating-Random Experiments (\textit{T-Rex}) selector~\cite{machkour2021terminating,machkour2022TRexGVS,machkour2023ScreenTRex}, which yields
\item FDR-controlled solutions while maximizing the number of selected (informative) variables and implicitly maximizing the explained (non-noise) variance.
\end{enumerate}

An implementation of the proposed \textit{T-Rex} PCA is available in the open source R package `TRexSelector' on CRAN~\cite{machkour2022TRexSelector}.

Organization: Section~\ref{sec: T-Rex Selector} revisits the \textit{T-Rex} selector. Section~\ref{sec: Proposed: T-Rex PCA} introduces the proposed \textit{T-Rex} PCA. In Sections~\ref{sec: Numerical Experiments} and~\ref{sec: Factor Analysis of SP 500 Stock Returns}, the results of numerical experiments and a factor analysis of S{\&}P $500$ stock returns are presented, respectively. Section~\ref{sec: Conclusion} concludes the paper.

\section{T-Rex Selector}
\label{sec: T-Rex Selector}
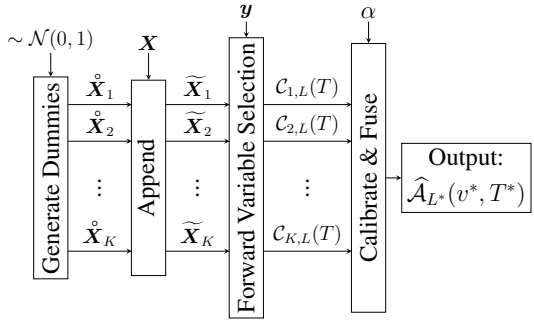
\begin{figure}[b]
\begin{center}
\scalebox{0.65}{
\begin{tikzpicture}[>=stealth]

  \coordinate (orig)   at (0,0);
  \coordinate (sample)   at (1,0.5);
  \coordinate (merge)   at (3,0.5);
  \coordinate (varSelect)   at (5,0.5);
  \coordinate (tFDR)   at (15.5,-1.1);
  \coordinate (fuse)   at (7.5,0.5);
  \coordinate (output)   at (9.5,0.5);
  
  \coordinate (between_scale_rank)   at (0.5,0.31);
  \coordinate (X_prime_to_tFDR_point)   at (0.5,6);
  \coordinate (X_prime_to_tFDR_point_point)   at (9,6);
  \coordinate (X_prime_to_merge_point)   at (0.5,-2.5);
  \coordinate (center_to_tFDR_point)   at (5.00,3.7);
  \coordinate (tFDR_to_fuse_point)   at (7.5,3.7);
  \coordinate (tFDR_to_sample_point)   at (4,5);

  \coordinate (Arrow_N_GenDummy)   at (1,3.06);
  \coordinate (Arrow_X_indVar)   at (3,3.06);
  \coordinate (Arrow_targetFDR_tFDR)   at (10,5.9);
  
   \coordinate (inference_Arrow)   at (15,-1.06);
   \coordinate (fuse_Arrow)   at (16.1,2.06);
  
  \coordinate (vdots1)   at (2.0,0.4);
  \coordinate (vdots2)   at (4.0,0.4);
  \coordinate (vdots3)   at (6.25,0.4);
  
  \coordinate (fuse_node)   at (15.00,0.5);
  
  \node[draw, minimum width=.7cm, minimum height=4cm, anchor=center , align=center] (C) at (sample) {\rotatebox{90}{\Large Generate Dummies}};
  \node[draw, minimum width=.7cm, minimum height=4cm, anchor=center, align=center] (D) at (merge) {\rotatebox{90}{\Large Append}};   
  \node[draw, minimum width=.7cm, minimum height=5.5cm, anchor=center, align=center] (E) at (varSelect) {\rotatebox{90}{\Large Forward Variable Selection}};
  \node[draw, minimum width=.7cm, minimum height=5.5cm, anchor=center, align=center] (H) at (fuse) {\rotatebox{90}{\Large Calibrate \& Fuse}};
  \node[draw, minimum width=2.5cm, minimum height=.7cm, anchor=center, align=center] (N) at (output) {\Large Output: \\[0.3em] \Large $\widehat{\mathcal{A}}_{L^{*}}(v^{*}, T^{*})$};
  \node (J) at (vdots1) {\Large $\vdots$};
  \node (K) at (vdots2) {\Large $\vdots$};
  \node (L) at (vdots3) {\Large $\vdots$};
  
  \draw[->] (Arrow_N_GenDummy) -- node[above, pos = 0.1]{\large $\sim\mathcal{N}(0, 1)$} ($(C.90)$); 
  \draw[->] (Arrow_X_indVar) -- node[above, pos = 0.1]{\large $\X$} ($(D.90)$); 
     
  \draw[->] ($(C.0) + (0,1.5)$) -- node[above]{\large $\XK_{1}$} ($(D.0) + (-0.7,1.5)$);
  \draw[->] ($(C.0) + (0,0.75)$) -- node[above]{\large $\XK_{2}$} ($(D.0) + (-0.7,0.75)$);
  \draw[->] ($(C.0) + (0,-1.5)$) -- node[above]{\large $\XK_{K}$} ($(D.0) + (-0.7,-1.5)$);
     
  \draw[->] ($(D.0) + (0,1.5)$) -- node[above]{\large $\XWK_{1}$} ($(E.0) + (-0.7,1.5)$);
  \draw[->] ($(D.0) + (0,0.75)$) -- node[above]{\large $\XWK_{2}$} ($(E.0) + (-0.7,0.75)$);
  \draw[->] ($(D.0) + (0,-1.5)$) -- node[above]{\large $\XWK_{K}$} ($(E.0) + (-0.7,-1.5)$);
     
  \draw[->] ($(E.0) + (0,1.5)$) -- node[above]{\large $\C_{1, L}(T)$} ($(H.0) + (-0.7,1.5)$);
  \draw[->] ($(E.0) + (0,0.75)$) -- node[above]{\large $\C_{2, L}(T)$} ($(H.0) + (-0.7,0.75)$);
  \draw[->] ($(E.0) + (0,-1.5)$) -- node[above]{\large $\C_{K, L}(T)$} ($(H.0) +  (-0.7,-1.5)$);
     
  \draw[->] (center_to_tFDR_point) -- node[above, pos = 0.1]{\large $\y$} ($(E.90)$); 
  \draw[->] (tFDR_to_fuse_point) -- node[above, pos = 0.1]{\Large $\alpha$} ($(H.90)$);
 
  \coordinate (between_varSelect_fuse1)   at ($(E.0) + (2.75,1.5)$);
  \coordinate (between_varSelect_fuse2)   at ($(E.0) + (2.75,0.75)$);
  \coordinate (between_varSelect_fuse3)   at ($(E.0) + (2.75,-1.5)$);
 
  \draw[->] (H) -- (N);
\end{tikzpicture}}
\end{center}
\setlength{\belowcaptionskip}{-8pt}
\caption{Simplified \textit{T-Rex} selector framework~\cite{machkour2021terminating,machkour2022TRexGVS}.}
\label{fig: T-Rex selector framework}
\end{figure}


The Terminating-Random Experiments (\textit{T-Rex}) selector is a fast and FDR-controlling variable selection framework for high-dimensional (and low-dimensional) data where $p > n$~\cite{machkour2021terminating}. As depicted in Fig.~\ref{fig: T-Rex selector framework}, it generates $K$ dummy matrices $\XK_{k} \in \mathbb{R}^{n \times L}$, $k = 1, \ldots, K$, containing $L$ standard normally distributed dummy predictors (see Theorem~2 of~\cite{machkour2021terminating}) that are appended to the original predictor matrix $\X$. It carries out $K$ independent random experiments by feeding the extended predictor matrices $\XWK_{k} = [ \X~\XK_{k} ]$ and the response vector $\y$ into a forward selection method, which yields $K$ candidate sets $\C_{k, L}(T)$, $k = 1, \ldots, K$. Each candidate set is the result of a random experiment that selects one variable at a time using a forward selection method, such as the \textit{LARS}~\cite{efron2004least} algorithm, \textit{Lasso}~\cite{tibshirani1996regression}, or \textit{elastic net}~\cite{zou2005regularization}, and terminates after $T$ dummies have been selected. The relative occurrence of each original variable $j \in \lbrace 1, \ldots, p \rbrace$ in the $K$ candidate sets is denoted by $\Phi_{T, L}(j)$. The final selected active set consists of all variables whose relative occurrences exceed a certain threshold $v \in [0.5, 1)$. That is, the selected set is given by
\begin{equation}
\widehat{\A}_{L^{*}}(v^{*}, T^{*}) \coloneqq \big\lbrace j : \Phi_{T^{*}, L^{*}}(j) > v^{*}  \big\rbrace,
\label{eq: selected active set}
\end{equation}
where the optimal triple $(T^{*}, L^{*}, v^{*}) \in \lbrace 1, \ldots, L^{*} \rbrace \times \mathbb{N}_{+} \times [0.5, 1)$ is determined by a calibration algorithm such that the FDR is controlled at a user defined target level $\alpha \in [0, 1]$ (i.e., $\FDR \leq \alpha$, see Theorem~1 of~\cite{machkour2021terminating}) while maximizing the number of selected variables and, thus, implicitly maximizing the true positive rate (TPR) (see Theorem 3 of~\cite{machkour2021terminating}). Given a selected active set $\widehat{\A}$ and the true active set $\A$, where $\widehat{\A}, \A \subseteq \lbrace 1, \ldots, p \rbrace$, the FDR and TPR are defined as (i) the expected fraction of false discoveries among all discoveries and (ii) the expected fraction of true discoveries among all true active variables, respectively. That is,
\begin{equation}
\FDR \coloneqq \mathbb{E} \bigg [ \dfrac{| \widehat{\A} \backslash \A|}{\max \lbrace 1, | \widehat{\A} | \rbrace} \bigg ] \text{~~\&~~} \TPR \coloneqq \mathbb{E} \bigg [ \dfrac{| \A  \cap \widehat{\A} |}{{\max \lbrace 1, | \A | \rbrace}} \bigg ],
\label{eq: FDR and TPR}
\end{equation}
where $| \cdot |$, $\backslash$, and $\cap$ are the cardinality, set exclusion, and intersection operators, respectively~\cite{machkour2021terminating}. Our goal is to control the FDR at low target levels while achieving a high TPR.

\section{Proposed: T-Rex PCA}
\label{sec: Proposed: T-Rex PCA}
In the following, the proposed \textit{T-Rex} PCA approach is explained and a comprehensive definition of the percentage of explained variance (PEV) for sparse PCA methods is presented.

\subsection{T-Rex PCA Algorithm}
\label{subsec: T-Rex PCA Algorithm}
We propose to leverage the \textit{T-Rex} selector to obtain FDR-controlled solutions of the formulation of sparse PCA as a collection of the $M$ \textit{elastic net} problems~\cite{zou2006sparse}, i.e.,
\begin{equation}
\left\lbrace
         \underset{\boldsymbol{\beta}_j \in \mathbb{R}^{p}}{\rm minimize}    
         \left\|   \z_{m} - \X \boldsymbol{\beta}_j  \right\|_{2}^{2}
         + \lambda_1 \left\|  \boldsymbol{\beta}_j  \right\|_1
         + \lambda_2 \left\|  \boldsymbol{\beta}_j \right\|_{2}^{2}
\right\rbrace_{m = 1}^{M}
\label{eq:Zou_SPCA}
\end{equation}
where $\lambda_{1}, \lambda_{2} > 0$ are tuning parameters and $\z_{m}$ is the plug-in estimate of the $m$th PC (i.e., the ordinary PC $\z_{m} = \X\bv_{m}$). The parameter $\lambda_{1}$ controls the sparsity level, while the ridge parameter $\lambda_{2}$ determines the strength of the variable grouping effect~\cite{zou2005regularization}.
\begin{algorithm}[b]
\caption{\textit{T-Rex PCA}.}
\begin{alglist}[leftmargin=0.6cm]
\itemsep-0.4em
\item \textbf{Input}: $\alpha$, $K$, $M$, $\X$, $\y$.
\label{Algorithm: T-Rex+SPCA - Step 1}
\item \textbf{Compute} the SVD of $\X$, i.e., $\X = \boldsymbol{U}\boldsymbol{D}\bV^{\top}$ and \textbf{determine} the ordinary PC matrix $\bZ = \big[ \z_{1}, \ldots, \z_{M} \big] = \boldsymbol{U}\boldsymbol{D}$ that contains the first $M \leq \min \lbrace n, p \rbrace$ ordinary PCs.
\label{Algorithm: T-Rex+SPCA - Step 2}
\item \textbf{For} $m = 1, \ldots, M$ \textbf{do}:
    \begin{enumerate}[leftmargin=0.4cm]
    \itemsep-0.4em
        \item[3.1.] \textbf{Run} the \textit{T-Rex} selector with
            \begin{enumerate}[leftmargin=0.3cm]
            \itemsep-0.4em
                \item[a.] the target FDR level $\alpha$,
                \item[b.] the extended predictor matrices $\XWK_{m, k} \coloneqq \big[ \X \,\, \XK_{m, k} \big]$, $k = 1, \ldots, K$, and
                \item[c.] the $m$th PC $\z_{m}$ as the common response for all $\XWK_{m, k}$, $k = 1, \ldots, K$.
            \end{enumerate}
        \item[3.2.] \textbf{Obtain} the FDR-controlled support of the $m$th loading vector $\widehat{\A}_{L_{m}^{*}}(v_{m}^{*}, T_{m}^{*})$.
        \item[3.3.] \textbf{Compute} the $m$th loading vector\\ $\hat{\bv}_{m} = \hatbbeta_{m, \rm Ridge} / \| \hatbbeta_{m, \rm Ridge} \|_{2}.$
        \vspace{0.3em}
         \item[3.4.] \textbf{Compute} the $m$th PC $\hat{\z}_{m} =\X_{\widehat{\A}_{m}}\hat{\bv}_{m}$.
    \end{enumerate}
\label{Algorithm: T-Rex+SPCA - Step 3}
\item \textbf{Output}: 
 \begin{enumerate}[leftmargin=0.4cm]
 \itemsep-0.4em
  \item[4.1.] \textit{T-Rex} supports $\widehat{\A}_{L_{m}^{*}}(v_{m}^{*}, T_{m}^{*})$, $m = 1, \ldots, M$, and
 \item[4.2.] \textit{T-Rex} PC matrix $\widehat{\bZ} = [ \widehat{\z}_{1}~\cdots~\widehat{\z}_{M}]$.
 \end{enumerate}
\label{Algorithm: T-Rex+SPCA - Step 4}
\end{alglist}
\label{Algorithm: T-Rex+SPCA}
\end{algorithm}

Our goal is to obtain FDR-controlled solutions of~\eqref{eq:Zou_SPCA} (i.e., $\lbrace \hatbbeta_{m} \rbrace_{m = 1}^{M}$) that provide a basis of (sparse) loading vectors for the dimension reduction. For this purpose, the ordinary PCs $\z_{m}$, $m = 1, \ldots, M$, serve as supervising response vectors within the \textit{T-Rex} selector (i.e., $\y = \z_{m}$ in Fig.~\ref{fig: T-Rex selector framework}) and we incorporate the \textit{elastic net} as the forward variable selector into the \textit{T-Rex} framework. This is achieved by reformulating the \textit{elastic net} as a \textit{Lasso}-type problem and solving it using the Terminating-LARS (\textit{T-LARS}) forward selection algorithm~\cite{machkour2022TRexGVS,machkour2022tlars}. This approach yields the sparse and FDR-controlled supports of the $M$ loading vectors, i.e., $\widehat{\A}_{L_{m}^{*}}(v_{m}^{*}, T_{m}^{*})$, $m = 1, \ldots, M$.

To convert the supports into loading vectors, we leverage the fact that the loading vectors can be linked to the ridge regression estimator~\cite{zou2006sparse}. That, in combination with the selected active set $\widehat{\A}_{m} \coloneqq \widehat{\A}_{L_{m}^{*}}(v_{m}^{*}, T_{m}^{*})$ as obtained by the \textit{T-Rex} selector, yields
\begin{equation}
\hat{\bv}_{m} = \hatbbeta_{m, \rm Ridge} / \| \hatbbeta_{m, \rm Ridge} \|_{2}, \quad m = 1, \ldots, M,
\label{eq: loading vectors ridge regression}
\end{equation}
where $\hatbbeta_{m, \rm Ridge} \coloneqq \arg \min_{\bbeta} \| \z_{m} - \X_{\widehat{\A}_{m}}\bbeta ||_{2}^{2} + \lambda_{2} \| \bbeta \|_{2}^{2}$ and $\X_{\widehat{\A}_{m}}$ contains only the predictors corresponding to $\widehat{\A}_{m}$. The \textit{T-Rex} PCs are then given by $\hat{\z}_{m} =\X_{\widehat{\A}_{m}}\hat{\bv}_{m}$, $m = 1, \ldots, M$.

Note that $\hat{\bv}_{m}$ is independent of $\lambda_{2}$ because of the scaling with the $\ell_{2}$-norm of $\hatbbeta_{m, \rm Ridge}$~\cite{zou2006sparse} and, therefore, we simply set $\lambda_{2} = 10^{-6}$. A major advantage of the \textit{T-Rex} selector framework is that when incorporating the \textit{elastic net} into it, the choice of $\lambda_{1}$ becomes obsolete, since the random experiments are terminated after $T^{*}$ dummies have entered the solution paths such that the FDR is controlled at the user-defined target level $\alpha$, which corresponds to choosing $\lambda_{1}$ for each random experiment such that an FDR-controlled selected active set $\widehat{\A}_{L_{m}^{*}}(v_{m}^{*}, T_{m}^{*})$ is obtained. The pseudocode of the proposed \textit{T-Rex} PCA is given in Algorithm~\ref{Algorithm: T-Rex+SPCA}.

The obtained FDR-controlled selected active sets can also be used to threshold the loading vectors of the ordinary PCA. Thus, in addition to the \textit{T-Rex} PCA, we also propose  the \textit{T-Rex} Thresholded PCA, which is obtained by thresholding each loading vector $\bv_{m}$ such that only the $| \widehat{\A}_{L_{m}^{*}}(v_{m}^{*}, T_{m}^{*}) |$ strongest loadings remain active (i.e., non-zero). The thresholded loading vector is then rescaled by its $\ell_{2}$-norm to ensure that $\| \hat{\bv}_{m} \|_{2} = 1$.

\subsection{Percentage of Explained Variance}
\label{subsec: Percentage of Explained Variance}
The explained variance (EV) in ordinary PCA is defined by $\tr(\widehat{\bZ}^{\top}\widehat{\bZ})$, where $\tr(\cdot)$ is the trace-operator. Since we are interested in the variance that corresponds to signal components, we define the percentage of explained variance (PEV) as follows:
\begin{defn}
Let $\widehat{\bV} = \widehat{\bV}_{\A} + \widehat{\bV}_{\A^{C}} \in \mathbb{R}^{p \times M}$, where $\widehat{\bV}_{\A}$ is the estimated loading matrix whose entries are set to zero except for the positions containing true active loadings and $\widehat{\bV}_{\A^{C}}$ is the estimated loading matrix whose true active loadings are set to zero. Then, $\widehat{\bZ} = \X \widehat{\bV} =   \X \widehat{\bV}_{\A} +  \X \widehat{\bV}_{\A^{C}} \eqqcolon \widehat{\bZ}_{\A} + \widehat{\bZ}_{\A^{C}}$ and the signal EV, mixed EV, and null EV are defined by
\begin{equation}
\EV \coloneqq \trace(\widehat{\bZ}^{\top} \widehat{\bZ}) =
\underbrace{
\trace(\widehat{\bZ}_{\A}^{\top} \widehat{\bZ}_{\A})
}_{\text{Signal EV}} + 
\underbrace{
2 \trace(\widehat{\bZ}_{\A}^{\top} \widehat{\bZ}_{\A^{C}})
}_{\text{Mixed EV}} + 
\underbrace{
\trace(\widehat{\bZ}_{\A^{C}}^{\top} \widehat{\bZ}_{\A^{C}})
}_{\text{Null EV}},
\label{PEV 1}
\end{equation}
and the PEV is defined by
\begin{equation}
\PEV \coloneqq \EV / (\text{Signal EV} + \text{Mixed EV}).
\label{PEV 4}
\end{equation}
\label{Definition: PEV}
\end{defn}
Our goal is to explain the signal and mixed EV with few PCs and sparse loadings to allow for interpretability of the obtained PCs. Non-sparse PCA methods or methods that do not provide accurate estimates of $\widehat{\bV}$ are prone to have a high null EV and, therefore, capture variance that merely corresponds to null (i.e., non-active) variables/loadings. In that case, the PEV in Definition~\ref{Definition: PEV} exceeds $100$\%, which indicates an inferior performance of the respective method. Moreover, since the orthogonality constraint in~\eqref{eq:generic_spca_optim} is dropped for sparse PCA methods, we replace the EV in Definition~\ref{Definition: PEV} by the adjusted EV that accounts for the lack of orthogonality of the loading vectors as suggested in~\cite{zou2006sparse}. The adjusted EV is defined by
\begin{equation}
\EV_{\rm adj} \coloneqq \sum\limits_{m = 1}^{M} r_{m,m}^{2},
\label{eq: adjusted EV}
\end{equation}
where $r_{m, m}$ is the $m$th diagonal element of  the upper triangular matrix $\mathbf{R}$ from the QR-decomposition of $\widehat{\bZ}$ (i.e., $\widehat{\bZ} = \mathbf{Q}\mathbf{R}$).

%
\begin{figure}[b]
  \centering
	\hspace*{-0.8em}
  \subfloat[First PC]{
  		\scalebox{1}{
  			\includegraphics[width=0.48\linewidth]{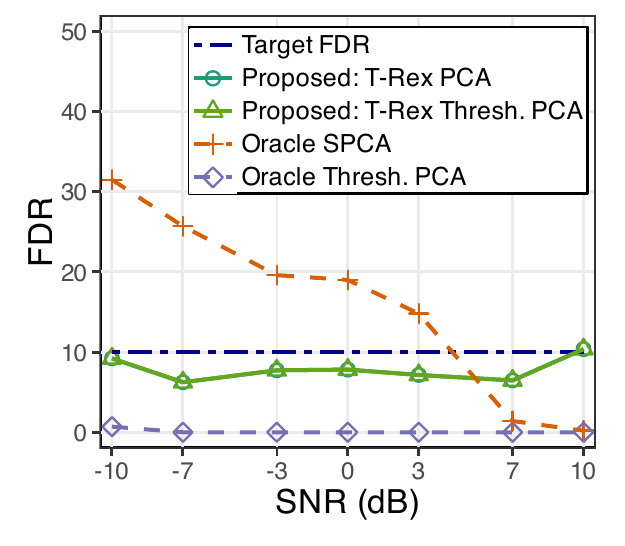}
  		}
   		\label{fig: FDR_vs_SNR_PC1}
   }
	\hspace*{-0.8em}
  \subfloat[First PC]{
  		\scalebox{1}{
  			\includegraphics[width=0.48\linewidth]{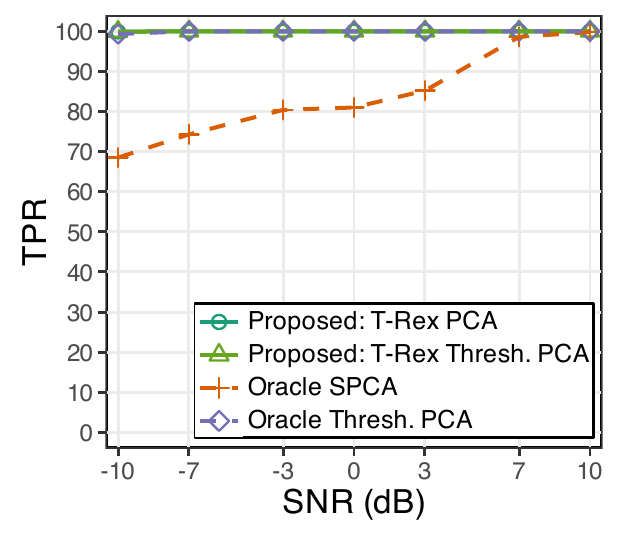}
  		}
   		\label{fig: TPR_vs_SNR_PC1}
   }
\setlength{\belowcaptionskip}{-8pt}
  \caption{For the first PC, the proposed \textit{T-Rex} PCA methods empirically control the FDR at a level of $10$\% while achieving an optimal TPR of $100$\% even at low SNRs. Only the infeasible oracle thresholded PCA achieves the same TPR at an FDR of almost zero. Except for high SNRs, the oracle SPCA is dominated by all other methods.}
  \label{fig: FDR_TPR_vs_SNR_PC1}
\end{figure}
%
\begin{figure*}[t]
  \centering
	\hspace*{-0.8em}
  \subfloat[]{
  		\scalebox{1}{
  			\includegraphics[width=0.24\linewidth]{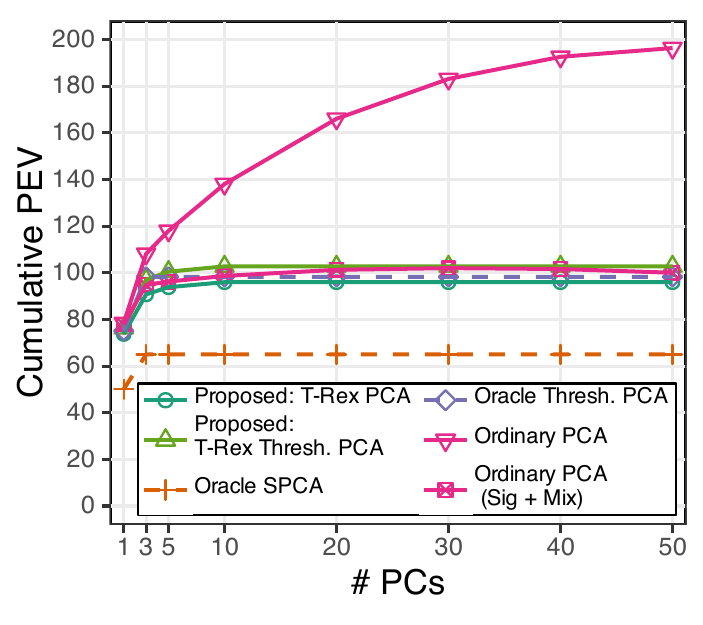}
  		}
   		\label{fig: cumPEV_vs_numPCs}
   }
	\hspace*{-0.8em}
  \subfloat[]{
  		\scalebox{1}{
  			\includegraphics[width=0.245\linewidth]{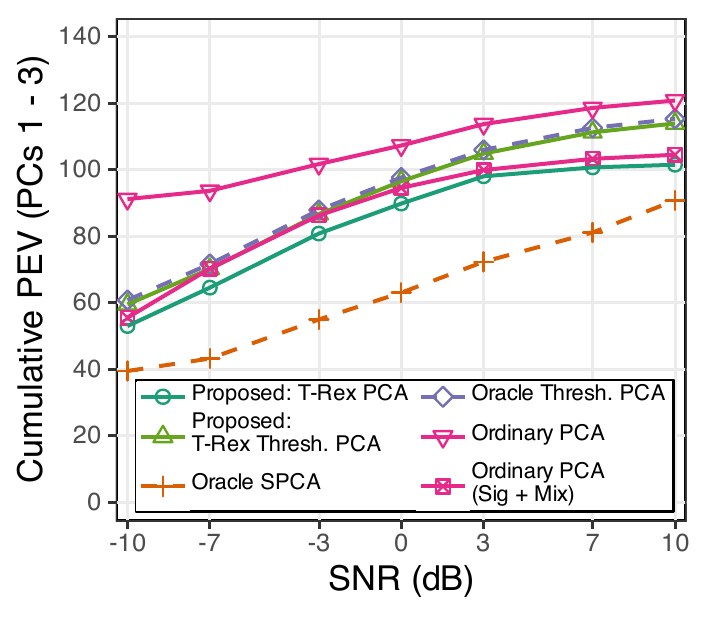}
  		}
   		\label{fig: cumPEV_vs_SNR}
   }
   %
   %
   %
   %
   %
   %
   	\hspace*{-0.8em}
  \subfloat[]{
  		\scalebox{1}{
  			\includegraphics[width=0.24\linewidth]{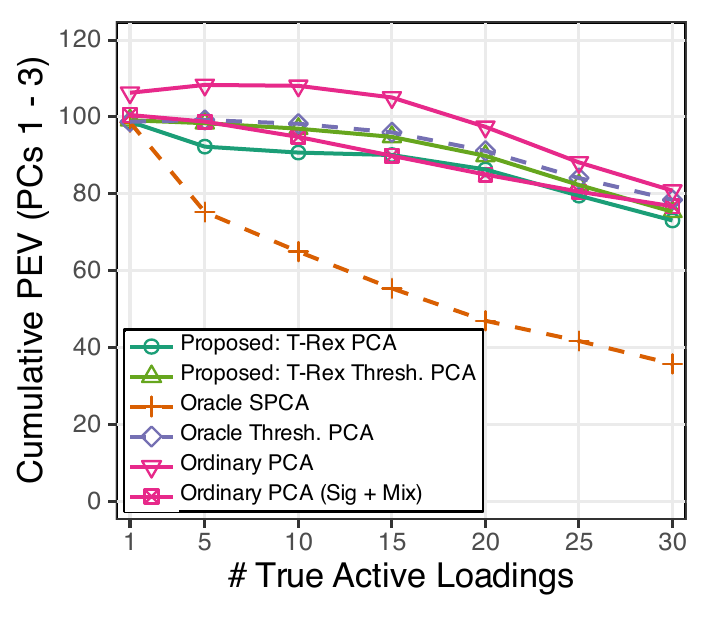}
  		}
   		\label{fig: cumPEV_vs_numTrueActiveLoadings}
   }
	\hspace*{-0.8em}
  \subfloat[]{
  		\scalebox{1}{
  			\includegraphics[width=0.24\linewidth]{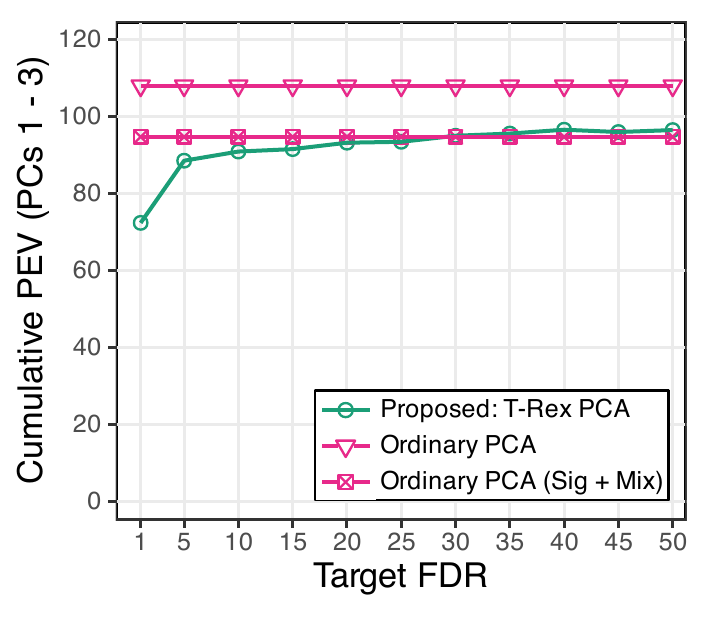}
  		}
   		\label{fig: cumPEV_vs_tFDR}
   }
   \setlength{\abovecaptionskip}{5pt}
\setlength{\belowcaptionskip}{-10pt}
  \caption{Cumulative percentage of explained variance (PEV): (a) - (c) As desired, the proposed \textit{T-Rex} PCA and \textit{T-Rex} Thresholded PCA require only very few PCs to explain the signal and mixed variance while not explaining any additional variance that is purely associated with null loadings. The oracle SPCA is outperformed by all other methods and the ordinary PCA explains all the variance in the data, including the variance that is merely associated with null loadings. (d) The cumulative PEV is not very sensitive with respect to the choice of the target FDR level for the  \textit{T-Rex} PCA, which allows the user to set almost any (preferably low) target FDR and still achieve a high cumulative PEV.
}
  \label{fig: cumPEV_vs_numPCs_SNR_numTrueActiveLoadings_tFDR}
\end{figure*}
%
\begin{figure*}[t]
  \centering
	\vspace*{0.2cm}
	\hspace*{-0.8em}
  \subfloat[No removed PC]{
  		\scalebox{1}{
  			\includegraphics[width=0.23\linewidth]{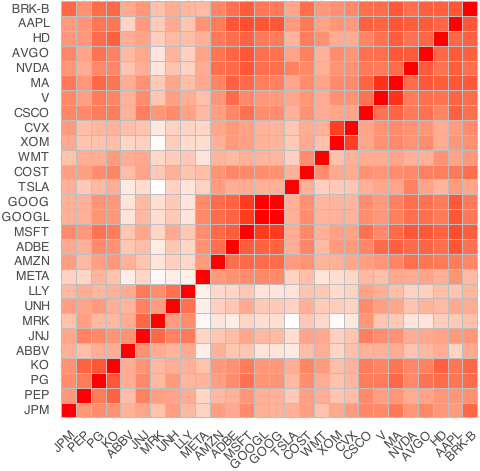}
  		}
   		\label{fig: corMat_noRemoved_PC_hcOrder}
   }
      %
   %
   %
   %
   %
   %
   	\hspace*{-0.8em}
  \subfloat[\small Ordinary PCA]{
  		\scalebox{1}{
  			\includegraphics[width=0.285\linewidth]{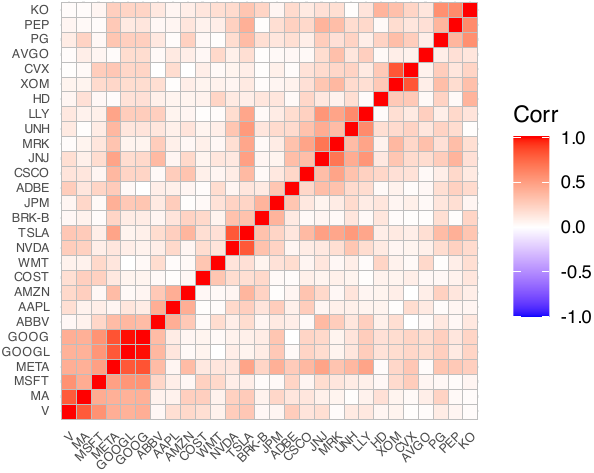}
  		}
   		\label{fig: corMat_3Removed_PC_hcOrder_pca}
   }
	\hspace*{-1.2em}
  \subfloat[Prop.: \textit{T-Rex} PCA]{
  		\scalebox{1}{
  			\includegraphics[width=0.23\linewidth]{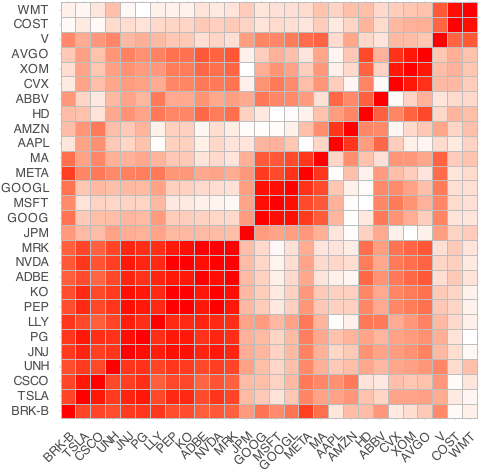}
  		}
   		\label{fig: corMat_3Removed_PC_hcOrder_trexSpca}
   }
   %
   %
   %
   %
   %
   %
	\hspace*{-1.2em}
  \subfloat[Prop.: \textit{T-Rex} Thresh. PCA]{
  		\scalebox{1}{
  			\includegraphics[width=0.23\linewidth]{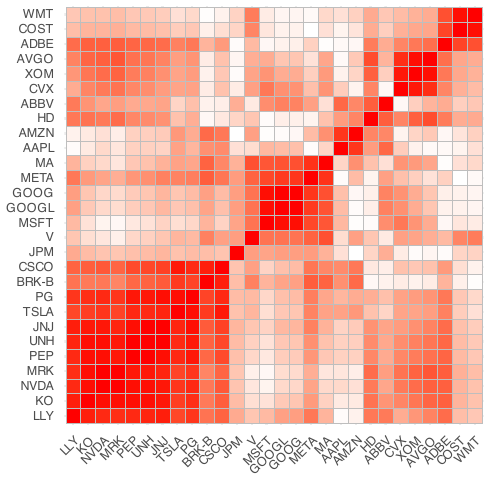}
  		}
   		\label{fig: corMat_3Removed_PC_hcOrder_threshPca}
   }
   \setlength{\abovecaptionskip}{5pt}
\setlength{\belowcaptionskip}{-10pt}
  \caption{Correlation matrices of the $28$ most influential stocks (according to their index weights) in the S{\&}P $500$ index.}
  \label{fig: corMat_SP500_trexSPCA_spca_threshPca_pca}
\end{figure*}
%

\section{Numerical Experiments}
\label{sec: Numerical Experiments}
We consider a high-dimensional data matrix $\X \in \mathbb{R}^{n \times p}$ with $n = 50$ samples, $p = 100$ variables, and centered columns that follows the sparse $M$-factor model
\begin{align}
\X &= \bZ\bV^{\top} + \boldsymbol{E} = [\z_{1} \cdots \z_{M}] [\bv_{1} \cdots \bv_{M}]^{T} + [\bepsilon_{1} \cdots \bepsilon_{p}] \\
&=
\begingroup
\renewcommand*{\arraystretch}{0.9}
\setlength\arraycolsep{1pt}
\begin{bmatrix}
z_{1, 1} & \cdots & z_{1, M} \\
z_{2, 1} & \cdots & z_{2, M} \\
\vdots &  & \vdots \\
z_{n, 1} & \cdots & z_{n, M} \\
\end{bmatrix}
\begin{bmatrix}
v_{1, 1} & \cdots & v_{p, 1} \\
v_{1, 2} & \cdots & v_{p, 2} \\
\vdots &  & \vdots \\
v_{1, M} & \cdots & v_{p, M} \\
\end{bmatrix}
+
\begin{bmatrix}
\epsilon_{1, 1} & \cdots & \epsilon_{1, p} \\
\epsilon_{2, 1} & \cdots & \epsilon_{2, p} \\
\vdots &  & \vdots \\
\epsilon_{n, 1} & \cdots & \epsilon_{n, p} \\
\end{bmatrix}
\endgroup,
\label{eq: factor model}
\end{align}
where $\z_{1}, \ldots, \z_{M}$ are Gaussian factors (i.e., $z_{i, m} \sim \mathcal{N}(0, \sigma_{m}^2)$), $\bv_{1}, \ldots, \bv_{M}$ are the corresponding sparse loading vectors of the factors (i.e., $v_{j, m} \in [0, 1]$), and $\bepsilon_{1}, \ldots, \bepsilon_{p}$ are Gaussian noise vectors (i.e., $\epsilon_{i, j} \sim \mathcal{N}(0, \sigma^2)$). We generate $M = 3$ factors with standard deviations $(\sigma_{1}, \sigma_{2}, \sigma_{3}) = (5, 3, 1)$. For each of the three factors, $p_{1}$ true active loadings are randomly selected among only the first $30$ out of $p = 100$ variables to simulate the more challenging case of overlapping loadings among the three factors. The values  of $p_{1}$ are varied over a range from $1$ to $30$. The values of the randomly selected loadings are set to $0.9$ (i.e., $v_{j, m} = 0.9$). The noise variance $\sigma^{2}$ is chosen such that the signal-to-noise ratio (SNR) is controlled over a range from $-10\dB$ to $+10\dB$. The SNR is defined by
\begin{equation}
\SNR \coloneqq 10 \log_{10} \big( \Var \big[\vectorize \big( \bZ\bV^{\top} \big) \big] \big/ \Var \big[ \vectorize \big( \boldsymbol{E} \big)\big] \big),
\end{equation}
where $\Var(\boldsymbol{a})$ and $\vectorize(\boldsymbol{A})$ denote the sample variance of a vector $\boldsymbol{a}$ and the vectorization operator that stacks the columns of a matrix $\boldsymbol{A}$ on top of each other, respectively. Finally, we set all simulation parameters that are not varied as follows: SNR = $0\dB$, $\alpha = 10$\% (target FDR level), $K = 20$ (number of \textit{T-Rex} random experiments; as suggested in~\cite{machkour2021terminating,machkour2022TRexGVS,machkour2023ScreenTRex,machkour2022TRexSelector}), $p_{1} = 5$ (number of true active loadings) in Fig.~\ref{fig: FDR_TPR_vs_SNR_PC1}, and $p_{1} = 10$ in Fig.~\ref{fig: cumPEV_vs_numPCs_SNR_numTrueActiveLoadings_tFDR}. The following three benchmark methods are considered:
\begin{enumerate}[label=\arabic*., ref=\arabic*, leftmargin=2em]
\itemsep-0.0001em
\item Ordinary (non-sparse) PCA.
\item The oracle thresholded PCA solution, which is obtained by thresholding each loading vector $\bv_{m}$ such that only the $p_{1}$ strongest loadings remain active (i.e., non-zero). The thresholded loading vector is then rescaled by its $\ell_{2}$-norm to ensure that $\| \hat{\bv}_{m} \|_{2} = 1$.
\item The oracle SPCA solution of~\eqref{eq:Zou_SPCA}, which is obtained by choosing the sparsity parameter $\lambda_{1}$ for each plug-in PC $\z_{m}$ such that only $p_{1}$ loadings remain active.
\end{enumerate}
Note that we are considering the best-case performances of the benchmark methods. In practice, however, only the proposed \textit{T-Rex} PCA methods are feasible without choosing any sparsity parameter. 

The results are averaged over $200$ Monte Carlo replications. A discussion is provided in the captions of Figures~\ref{fig: FDR_TPR_vs_SNR_PC1} and~\ref{fig: cumPEV_vs_numPCs_SNR_numTrueActiveLoadings_tFDR}.

\section{Factor Analysis of S{\&}P 500 Stock Returns}
\label{sec: Factor Analysis of SP 500 Stock Returns}
Understanding the interdependencies among stocks in an index such as the S{\&}P $500$ index is crucial for the analysis of portfolios. However, computing a simple sample correlation matrix does not allow to assess the fine interdependencies among stocks. The reason is that all stocks in the S{\&}P $500$ index are part of the same market and, therefore, are obscured by strong statistical market factors~\cite{ruppert2011statistics,avellaneda2010statistical}. Our goal in this application is, therefore, to determine the strongest common factors and remove them from the data (which leaves us with the idiosyncratic component) to reveal the fine interdependencies among the stocks. That is, we compute $\widehat{\X} = \widehat{\bZ}^{\prime} \widehat{\bV}^{\prime \top}$, where $\widehat{\bZ}^{\prime}$ and $\widehat{\bV}^{\prime}$ are copies of $\widehat{\bZ}$ and $\widehat{\bV}$, respectively, except that the first three columns (i.e., the first three PCs) are removed. For this purpose, we consider the returns of the stocks that constitute the S{\&}P $500$ index in the three month period from $2022$-$10$-$1$ to $2022$-$12$-$31$. Hence, the matrix $\X = [\x_{1} \cdots \x_{p}] \in \mathbb{R}^{n \times p}$ contains $n$ daily returns of $p$ stocks $\x_{j} = [ x_{1, j} \cdots x_{n, j} ]^{\top} \in \mathbb{R}^{n}$, $j = 1, \ldots, p$. The returns of the $j$th stock are given by
\begin{equation}
x_{i, j} = ( \price_{i, j} - \price_{i - 1, j}) / \price_{i - 1, j}, \, i = 2, \ldots, n,
\label{eq: daily asset returns}
\end{equation}
where $\price_{i, j}$ is the closing price of the $j$th stock on day $i$.

Fig.~\ref{fig: corMat_SP500_trexSPCA_spca_threshPca_pca} presents the correlation matrices of the $28$ most influential stocks (i.e., stocks with index weight larger than $0.6$\%) in the S{\&}P $500$ index. In order to visually distinguish groups of highly associated stocks, the correlation matrices are reordered using complete linkage hierarchical clustering. Even after reordering, the correlation matrix that corresponds to no removed PCs barely reveals any groups of stocks. The ordinary PCA removes too much variance and, therefore, does not allow to distinguish groups of highly correlated stocks. In contrast, after removing the first three PCs, the proposed methods (i.e., \textit{T-Rex} PCA and \textit{T-Rex} Thresholded PCA at a target FDR level of $10$\%) reveal that there exist meaningful groups of highly correlated stocks that are not explained by the three leading PCs but by the idiosyncratic component. Since the oracle SPCA is infeasible in this real world example, it is omitted. The results indicate meaningful relationships among stocks from different industries. However, a detailed interpretation of the results from a portfolio design perspective goes beyond the scope of this paper.

\section{Conclusion}
\label{sec: Conclusion}
The proposed \textit{T-Rex} PCA and \textit{T-Rex} Thresholded PCA perform the double duty of dimension reduction and variable selection while controlling the FDR of the sparse loading vectors. They require no tuning of any sparsity parameters and are capable of explaining the signal variance in the data with few PCs, which allows for meaningful interpretations of the PCs. The proposed methods showed a promising performance in simulated data and proved to be useful for revealing the interdependencies among stocks from the S{\&}P $500$ index.

\vfill\pagebreak

\bibliographystyle{IEEEbib}
\bibliography{refs}

\typeout{get arXiv to do 4 passes: Label(s) may have changed. Rerun}

\end{document}